\documentclass{article}
\usepackage{spconf,amsmath,amsthm,amsfonts,graphicx,url,booktabs,multirow,subcaption,bm,bbm,breqn,xcolor}

\def\vh{{\bm{h}}}

\def\vx{{\bm{x}}}

\def\vth{\Tilde{\vh}}

\def\gN{{\mathcal{N}}}
\def\mI{{\bm{I}}}

\def\vzero{{\bm{0}}}

\title{Single-View Height Estimation with \\
Conditional Diffusion Probabilistic Models}

\name{Isaac Corley \qquad Peyman Najafirad}

\address{Secure Artificial Intelligence Laboratory for Autonomy (AILA) \\
The University of Texas at San Antonio, Texas, USA \\
\{isaac.corley,  peyman.najafirad\}@utsa.edu}

\begin{document}
\maketitle

\begin{abstract}
Digital Surface Models (DSM) offer a wealth of height information for understanding the Earth's surface as well as monitoring the existence or change in natural and man-made structures. Classical height estimation requires multi-view geospatial imagery or LiDAR point clouds which can be expensive to acquire. Single-view height estimation using neural network based models shows promise however it can struggle with reconstructing high resolution features. The latest advancements in diffusion models for high resolution image synthesis and editing have yet to be utilized for remote sensing imagery, particularly height estimation. Our approach involves training a generative diffusion model to learn the joint distribution of optical and DSM images across both domains as a Markov chain. This is accomplished by minimizing a denoising score matching objective while being conditioned on the source image to generate realistic high resolution 3D surfaces. In this paper we experiment with conditional denoising diffusion probabilistic models (DDPM) for height estimation from a single remotely sensed image and show promising results on the Vaihingen benchmark dataset.
\end{abstract}

\begin{keywords}
diffusion models, digital surface models, remote sensing, geospatial, height estimation
\end{keywords}

\section{Introduction}
\label{sec:intro}
\begin{figure}[ht!]
  \centering
  \centerline{\includegraphics[width=0.8\linewidth]{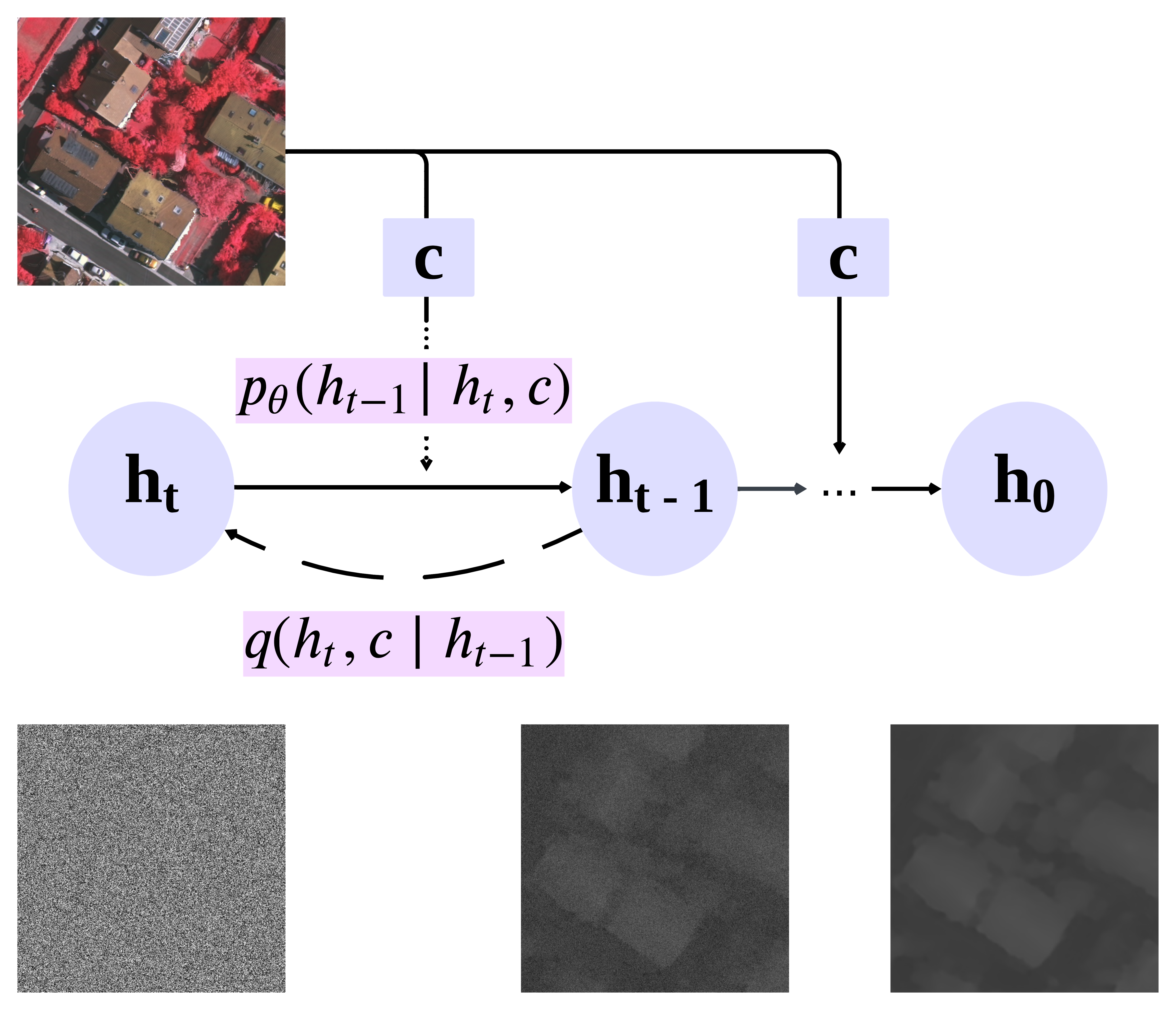}}
\caption{An overview of the Image Conditioned Denoising Diffusion Probabilistic Models (DDPM) architecture for single-view height estimation of remotely sensed imagery. Our architecture is a modified form of the DDPM processes which conditions on each step with the source image, $c$ to generate high resolution predictions.}
\label{fig:arch}
\end{figure}

Height estimation and digital surface model generation from aerial and/or satellite imagery has numerous applications including monitoring change of the Earth~\cite{qin20163d}, humanitarian and disaster response (HADR)~\cite{tu2016automatic}, and urban planning~\cite{beumier2016digital}. This data, in combination with the remotely sensed imagery, can be used to generate 3D approximations of models of scenes on the Earth. However, in comparison to natural images, remotely sensed imagery is inherently complex due to variations in spatial resolution, off-nadir acquisitions, seasonal changes, clouds, etc. These challenges require the use of powerful models to learn through the noise. Recently, diffusion models have grown to the forefront of academic research for the dramatic improvements in high resolution image synthesis and generation. The experiments conducted in this paper seek to explore the intersection of diffusion models and remotely sensed imagery, particularly for the task of height estimation from a single image.

To our knowledge, no other methods have trained a Denoising Diffusion Probabilistic Models (DDPM) generative model to learn the joint distribution of optical and Digital Surface Model (DSM) remotely sensed imagery. Our contributions can be described as following:

\begin{figure*}[ht!]
  \centering
  \centerline{\includegraphics[width=0.8\textwidth]{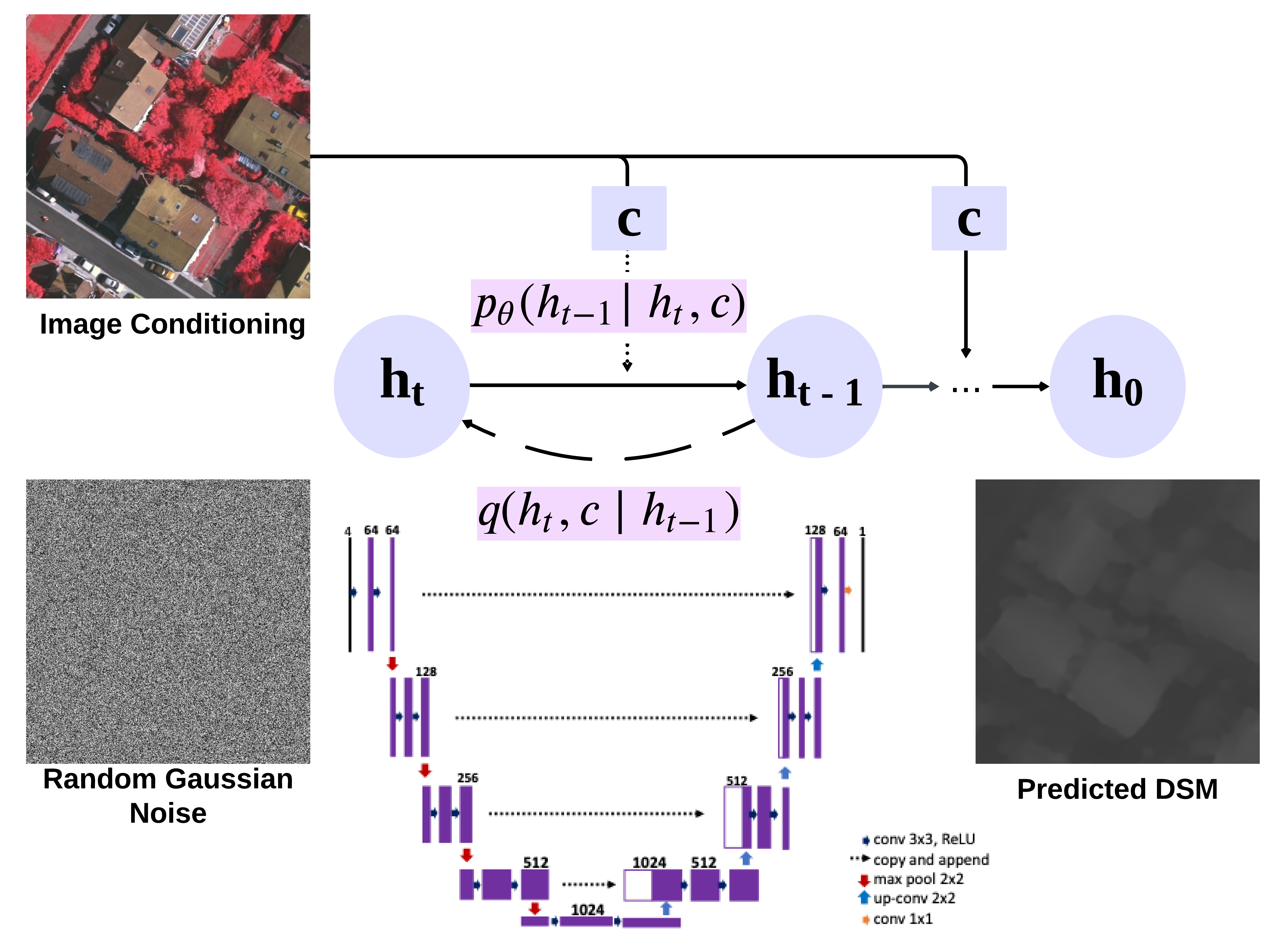}}
\caption{An overview of the Image Conditioned DDPM pipeline with the U-Net architecture. During inference we sample a random white noise image $\vth_{T}$ and iteratively pass the noise image conditioned with the single source image, $c$, through the U-Net model to generate $\vth_{t}$ over some number of time steps, $t$, until we reach the final denoised height output, $\vth_{0}$, which we train using the $L_{2}$ loss to approximate the ground truth DSM, $\vh$.}
\label{fig:arch-unet}
\end{figure*}

\begin{itemize}
    \item \textit{Image Conditioned Diffusion Models} - We use a variation of Diffusion Models which conditions each step with source image to guide the generative process.
    \item \textit{State-of-the-art Single-View Height Estimation Performance} - Our novel use of conditional diffusion models for single-view height estimation method quantitatively and qualitatively outperforms previous works.
\end{itemize}

\subsection{Diffusion Models}
DDPMs~\cite{ho2020denoising} are a class of latent variable generative models inspired by Markovian dynamics. Recently, this method has overtaken other generative methods such as Generative Adversarial Networks (GAN)~\cite{goodfellow2020generative}, Normalizing Flows~\cite{rezende2015variational}, and Variational Autoencoders (VAE)~\cite{kingma2013auto}, resulting in state-of-the-art performance for several research topics including image super resolution~\cite{saharia2022palette}, high-resolution image synthesis~\cite{saharia2022photorealistic}, and text-to-image generation~\cite{rombach2021highresolution}. Diffusion models are tractable and stable to train and sample from in comparison to said generative methods and appear to have much unexplored potential for adoption in other fields. With interest in diffusion models increasing, remote sensing research has begun to investigate the efficacy of these methods for geospatial and remotely sensed imagery and data~\cite{bandara2022ddpm, liu2022diffusion, kolbeinsson2022multi}. However, diffusion models have yet to be explored for the task of single-view height estimation.

\subsection{Single-View Height Estimation}
Classical remote sensing height estimation relies on either multiple imagery sources for use with stereo or multi-view reconstruction and photogrammetry techniques or LiDAR data. However, the data acquisition and processing for these methods can become costly. As a result, deep learning based methods using only a single-view for height estimation have become popular for obtaining inexpensive estimates of pixelwise height in remotely sensed imagery. With that said, estimating height from a single viewpoint is incredibly due to the lack of information. This has resulted in an explosion in research utilizing deep learning techniques to learn from optical imagery and digital surface models, including convolutional neural networks (CNNs)~\cite{mou2018im2height,liu2020im2elevation} and GANS~\cite{ghamisi2018img2dsm}, and self-supervised learning~\cite{corley2022supervising}. While many methods perform multitask learning to learn both pixelwise height and land cover jointly, in this work we focus on purely methods for height estimation.

\section{Methods}
\label{sec:methods}
In the case of single-view height estimation we seek to optimize $\min_{\theta} \sum_{i=1}^{\infty}(h_{i} - \hat{h_{i}})$, by training a black box model $f_\theta$, in this case a U-Net \cite{ronneberger2015u} architecture, which takes in an optical image, $c_{i}$, and produces a height estimate, $\hat{h_{i}} = f_\theta(c_{i})$.

\subsection{Denoising Diffusion Probabilistic Models}
\label{sec:methods-ddpm}

For applying DDPMs for single-view height estimation we use variants of the typical diffusion forward process, $q$, and reverse process, $p$. Given a ground truth DSM $\vh$, we perform the forward process by iteratively adding Gaussian noise of increasing scale over $T$ steps to $\vh$ until we reach a final white noise state $\vth_{T}$. Given $\vth_{T}$, we seek to generate a height estimate through the reverse process of training a model, $f_\theta$, to iteratively denoise the input given a noise level indicator, $\gamma$, until we reach the final denoised approximation of the ground truth, $\vh_{0}$. Additionally, we condition the reverse process to generate the height estimate of a given source optical image, $c$, by concatenation at each time step, $t$, with the height estimate, $\vth_{t}$. The resulting loss function which we optimize the network for single-view height estimation using the $L_{p}$ norm at $p=2$ in the equation is provided in the algorithm below.

In this work we specifically utilize diffusion implicit models (DDIMs)~\cite{song2020denoising} which are a variation of DDPMs that improve the inference sampling rate. To summarize, the DDPM generative forward process is converted from a Markovian to an implicit probabilistic model by setting $\sigma_t=0$ in Equation \ref{eq:sample-eq-gen} below, which makes the process deterministic.

Defining $\alpha_0=1$ and $\epsilon_t \sim \gN(\vzero, \mI)$ is to be standard Gaussian, $p_\theta(\vx_{1:T})$, one can generate a sample $\vx_{t-1}$ from a sample $\vx_{t}$ given:
\begin{dmath}
    \vx_{t-1} = \sqrt{\alpha_{t-1}} \left(\frac{\vx_t - \sqrt{1 - \alpha_t} \epsilon_\theta^{(t)}(\vx_t)}{\sqrt{\alpha_t}}\right) \vx_0 + \sqrt{1 - \alpha_{t-1} - \sigma_t^2} \cdot \epsilon_\theta^{(t)}(\vx_t) + \sigma_t \epsilon_t \label{eq:sample-eq-gen}
\end{dmath}

\section{Experiments}
\label{sec:experiments}
\subsection{Dataset}
The Vaihingen dataset~\cite{rottensteiner2012isprs} is part of the ISPRS benchmark on urban object classification and 3D building reconstruction. The dataset contains 33 patches of various sizes of pairs of optical imagery and digital surface models extracted from a larger orthomosaic. The imagery contains 3 bands (near-infrared (NIR), red, and green). The ground sampling distance or spatial resolution of the imagery and DSMs is 9cm per pixel. The DSMs were generated using the Trimble INPHO 5.3 software using classical dense image matching methods. We use the provided benchmark train and test set splits throughout our experiments.

\subsection{Experimental Details and Hyperparameters}
We utilize a NVIDIA DGX server and a single NVIDIA A100 node with 80GB GPU RAM for training and inference sampling. We normalize the inputs and outputs to the range $[-1, 1]$.

\begin{description}
    \item[Training] During training we random crop 512x512 chips from the train set orthomosaics and DSMs. We perform random augmentations to all data including horizontal/vertical flipping, random rotation, as well as apply random color jitter and gaussian blurring to the optical imagery only. We train using $T=1000$ timesteps and a sigmoid variance schedule. We train for 100k iterations, a batch size of 8, mixed precision training, gradient accumulation every 2 steps, and use the Adam~\cite{kingma2014adam} optimizer with a learning rate of $\alpha=1e-3$.

    \item[Inference] During inference we use a chip size of 1024x1024. As mentioned in Section \ref{sec:methods-ddpm}, we utilize DDIM with $T=500$ and $\eta=0$ for faster sampling. Additionally, we utilize FP16 mixed precision during sampling to reduce the memory usage.
\end{description}

\begin{table}[h]
\centering
\begin{tabular}{@{}ccc@{}}
\toprule
\textbf{Method} & \multicolumn{1}{c}{\textbf{MAE (m) $\downarrow$}} & \multicolumn{1}{c}{\textbf{RMSE (m) $\downarrow$}} \\
\midrule
IM2HEIGHT~\cite{mou2018im2height} & 1.485 & 2.253 \\
D3Net~\cite{carvalho2018regression} & 1.314 & 2.123 \\
Amirkolaee et al.~\cite{amirkolaee2019height} & 1.487 & 2.197 \\
3DBR~\cite{alidoost20192d} & 1.379 & 2.074 \\
IM2ELEVATION~\cite{liu2020im2elevation} & 1.226 & 1.882 \\
PLNet~\cite{xing2021gated} & 1.178 & 1.775 \\
DDPM (Ours) & \textbf{1.093} & \textbf{1.760} \\ 
\bottomrule
\end{tabular}
\caption{
Experimental results on the Vaihingen dataset with comparisons to other single-task height estimation methods. Benchmark results for other methods are referenced from~\cite{xing2022sce}. Best results are marked in bold.}
\label{tab:results}
\end{table}

\section{Discussion and Future Work}
\label{sec:discussion}
\begin{figure*}[ht!]
  \centering
  \centerline{\includegraphics[width=0.65\textwidth]{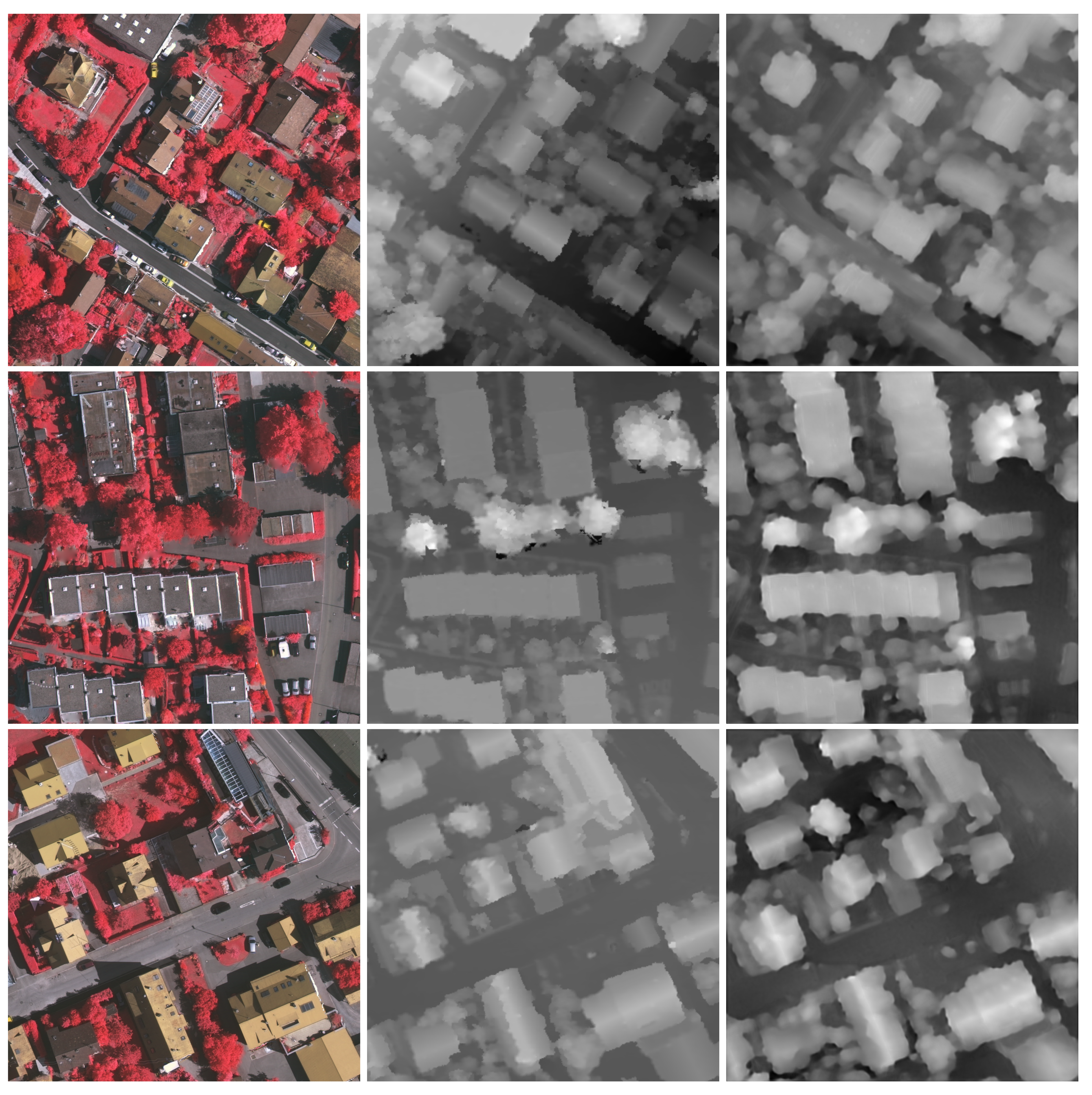}}
\caption{Random Vaihingen Test Set Samples. From left to right: Optical image, Ground Truth, DDPM Predictions}
\label{fig:results}
\end{figure*}

The experimental results in Table \ref{tab:results} provide evidence that training diffusion based models for single-view height estimation outperforms other previous benchmark methods. It is notable to metniond that ground truth DSMs used in many benchmarks are noisy, most notably around the edges of buildings or tree crowns, because they were generated using traditional methods. As seen in Figure \ref{fig:results}, the DDPM output height estimates tend to learn to produce less noisy outputs and instead smooth these edges, which could lead to better downstream perform for generating 3D models of cities and roof wireframes.

While our experiments only explore comparisons of aerial and satellite imagery based models, we note that investigation of Vision Transformers (ViT)~\cite{dosovitskiy2020image} based models would be an interesting future work to replace the U-Net backbone in the DDPM pipeline. ViTs have shown improved performance in segmentation and change detection applications in comparison to fully-convolutional based architectures~\cite{chen2021remote}.

\section{Conclusion}
\label{sec:conclusion}
In this paper we explore cutting edge DDPMs for estimating pixelwise height to generate high resolution digital surface models from remotely sensed imagery. We find that DDPMs outperform non diffusion based models and show promise for producing accurate surface estimates of the Earth which has many downstream applications with the potential to benefit the monitoring of manmade and natural change and 3D generation of cities. We hope that this work will inspire exploration of diffusion models in the remote sensing field.

\bibliographystyle{IEEEbib}
\bibliography{refs}

\end{document}